\documentclass[10pt,twocolumn,letterpaper]{article}

\usepackage{cvpr}
\usepackage{times}
\usepackage{epsfig}
\usepackage{graphicx}
\usepackage{amsmath}
\usepackage{amssymb}

\usepackage{bm}
\usepackage{enumitem}
\usepackage{booktabs}
\usepackage{multirow}
\usepackage{graphicx}
\usepackage{caption}
\usepackage{subcaption}
\usepackage[ruled,linesnumbered]{algorithm2e}
\usepackage{rotating}
\usepackage{optidef}
\usepackage[table]{xcolor}
\setlist[itemize]{leftmargin=4mm}

\usepackage[pagebackref=true,breaklinks=true,colorlinks,bookmarks=false]{hyperref}

\DeclareMathOperator*{\argmin}{arg\,min}

\cvprfinalcopy 

\def\cvprPaperID{1625} 

\ifcvprfinal\fi

\begin{document}

\title{Autolabeling 3D Objects with Differentiable Rendering of SDF Shape Priors}

\author{Sergey Zakharov$^*$\\
Technical University of Munich \\
{\tt\small sergey.zakharov@tum.de}
\and
Wadim Kehl$^*$, Arjun Bhargava, Adrien Gaidon \\
Toyota Research Institute\\
{\tt\small {firstname.lastname}@tri.global}
}

\maketitle


\begin{abstract}
   We present an automatic annotation pipeline to recover 9D cuboids and 3D shapes from pre-trained off-the-shelf 2D detectors and sparse LIDAR data. Our autolabeling method solves an ill-posed inverse problem by considering learned shape priors and optimizing geometric and physical parameters. To address this challenging problem, we apply a novel differentiable shape renderer to signed distance fields (SDF), leveraged together with normalized object coordinate spaces (NOCS). Initially trained on synthetic data to predict shape and coordinates, our method uses these predictions for projective and geometric alignment over real samples. Moreover, we also propose a curriculum learning strategy, iteratively retraining on samples of increasing difficulty in subsequent self-improving annotation rounds. Our experiments on the KITTI3D dataset show that we can recover a substantial amount of accurate cuboids, and that these autolabels can be used to train 3D vehicle detectors with state-of-the-art results. The code is available at \url{http://github.com/TRI-ML/sdflabel}.
\end{abstract}

\let\thefootnote\relax\footnote{$^*$ First co-authorship. This work resulted from an internship at TRI.}

\section{Introduction}

\begin{figure}[t]
	\centering
	\includegraphics[width=1\linewidth]{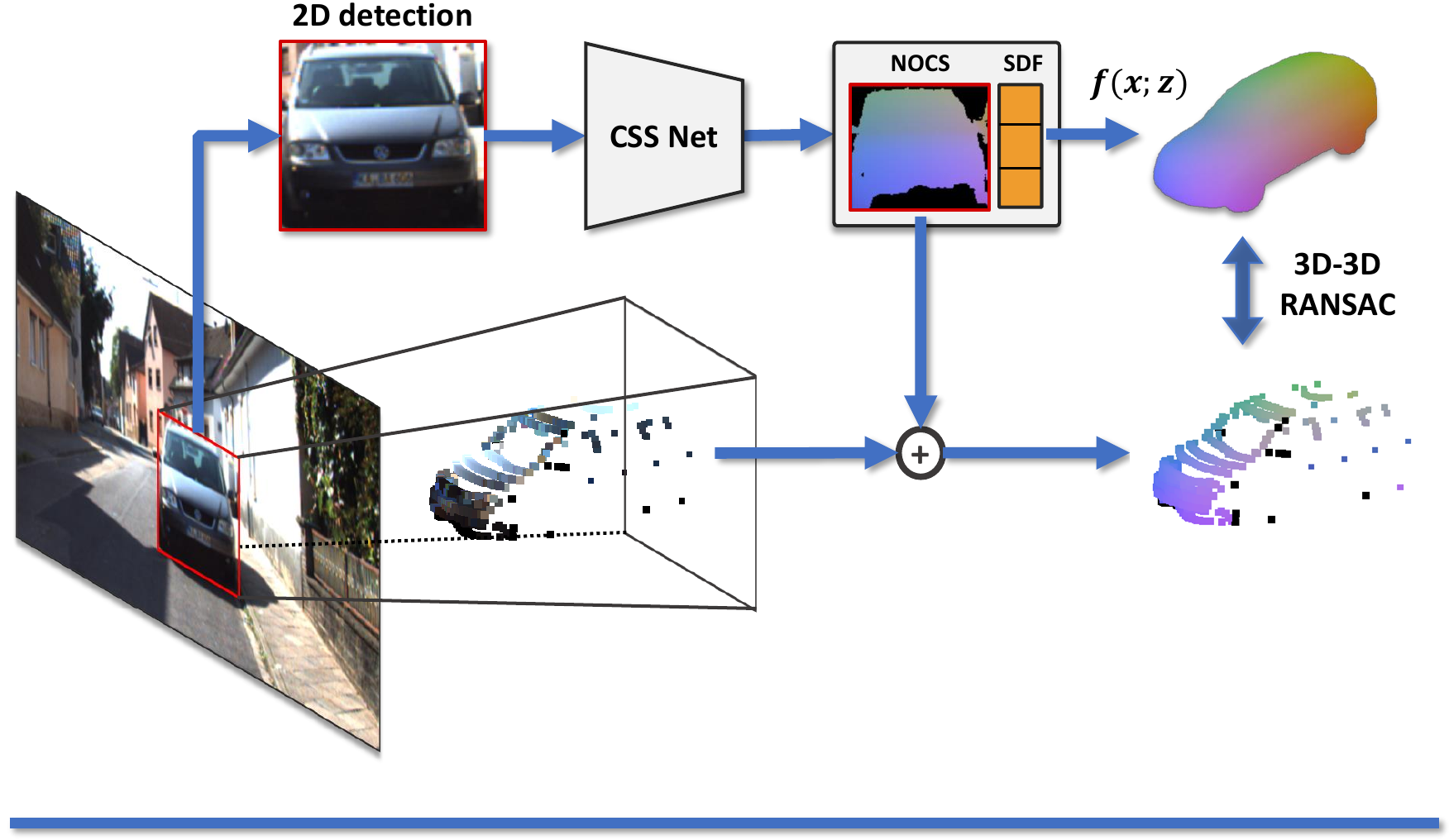}
	\includegraphics[width=1\linewidth, height=4.2cm]{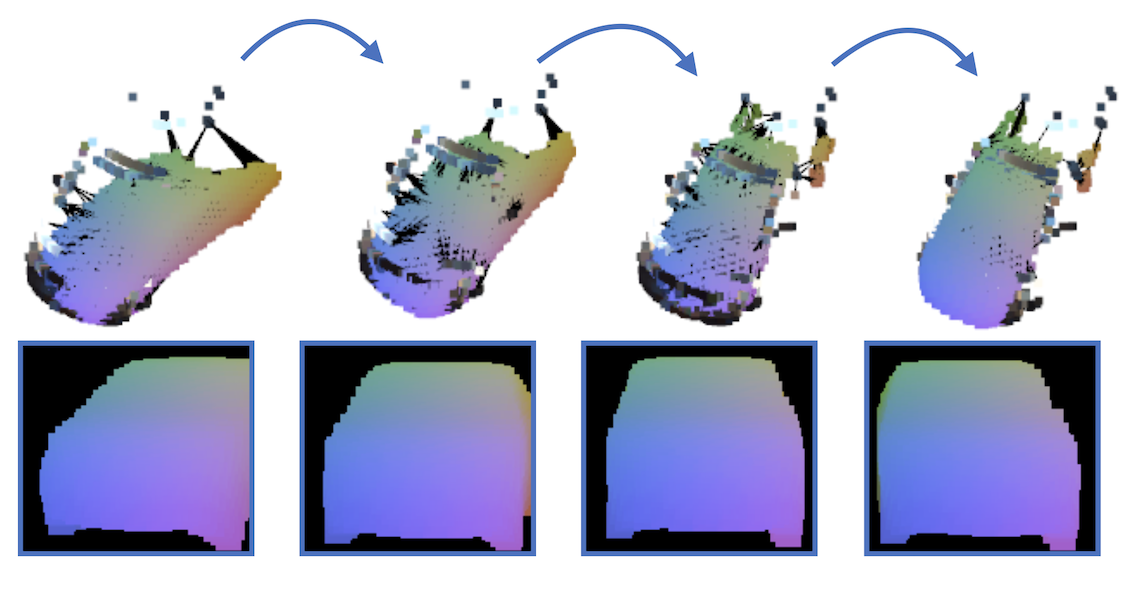}
	\caption{Our pipeline for 3D object autolabeling. Top: off-the-shelf 2D detections are fed into our Coordinate Shape Space (CSS) network to predict surface coordinates and a shape vector. We backproject the coordinates to LIDAR in the camera frustum and decode the shape vector into an object model. Then, we establish 3D-3D correspondences between the scene and model to estimate an initial affine transformation. Bottom: We iteratively refine the estimate via differentiable geometric and visual alignment.}
	\label{fig:intro}  
\end{figure}

Deep learning methods require large labeled datasets to achieve state-of-the-art performance. Concerning object detection for automated driving, 3D cuboids are preferred among other annotation types as they allow appropriately reasoning over all nine degrees of freedom (instance location, orientation, and metric extent).
However, obtaining a sufficient amount of labels to train 3D object detectors is laborious and costly, as it mostly relies on involving a large number of human annotators.
Existing approaches for scaling up annotation pipelines include the usage of better tooling, active learning, or a combination thereof \cite{Lee2018, Huang2018, Wang2019, Ling2019, Chen2019}. Such approaches often rely on heuristics and require human effort to correct the outcomes of semi-automatic labeling, specifically for difficult edge cases.

Alternatively, we propose a novel approach relying on differentiable rendering of shape priors to recover metric scale, pose, and shape of vehicles in the wild. Our 3D autolabeling pipeline requires only 2D detections (bounding boxes or instance masks) and sparse point clouds (ubiquitous in 3D robotic contexts). Detections themselves are produced using off-the-shelf 2D detectors. We demonstrate that differentiable visual alignment, also referred to as~\textit{``analysis-by-synthesis''}~\cite{yuille2006vision} or \textit{``render-and-compare''}~\cite{Kundu2018}, is a powerful approach towards autolabeling for the purpose of autonomous driving. 

The present study introduces three novel contributions. First, we formulate the notion of a \textbf{Coordinate Shape Space (CSS)}, which combines Normalized Object Coordinates (NOCS)~\cite{Wang2019NOCS} with the DeepSDF framework~\cite{park2019deepsdf}. This allows to reliably set object shapes into correspondence to facilitate \textit{deformable shape matching}. Second, we present a way to differentiate DeepSDF with respect to its surface, thereby introducing a novel \textbf{differentiable SDF renderer} for comparative scene analysis over a defined shape space. The third contribution is a \textbf{curriculum learning-based autolabeling pipeline} of driving scenes. Figure~\ref{fig:intro} presents an example optimization on the KITTI3D dataset~\cite{Geiger2012}. 

Our pipeline starts with a CSS network, which is trained to predict 2D NOCS maps as well as shape vectors from image patches. To bootstrap an initial version, we train the network on synthetic data, for which we can easily obtain ground truth NOCS and shape vector targets, and apply augmentations to minimize the sim2real domain gap. 
Our autolabeling loop includes the following steps: 1) leveraging 2D detections to localize instances; 2) running the CSS network on an extracted patch; 3) reprojecting NOCS into the scene using LIDAR, 4) decoding an object model from the shape space; 5) computing an approximate pose using 3D-3D correspondences; and 6) running projective and geometric alignment for refinement. 
After processing all images, we collect the recovered \textit{autolabels} and retrain our CSS prediction network to gradually expand it into the new domain. Then, we repeat this process to achieve better CSS predictions and, consequently, better autolabels. To avoid drifting due to noisy autolabels, we employ a curriculum that is focused on easy samples first and increases the difficulty in each loop.    


In summary, \textbf{our main contributions} are as follows: (i) a novel, fully-differentiable renderer for signed distance fields that can traverse smooth shape spaces; (ii) a mixed synthetic/real curriculum framework that learns to predict shape and object coordinates on image patches; and (iii) a multi-modal optimization pipeline combining differentiable alignment over vision and geometry.  
We evaluate our approach on the KITTI3D dataset~\cite{Geiger2012} and show that our method can be used to accurately recover metric cuboids with structural, differentiable priors. Furthermore, we demonstrate that such cuboids can be leveraged to train efficient 3D object detectors.

\section{Related Work}

In recent years, assisted labeling has gained growing attention as the increasing amount of data hinders the ability to label it manually. In \cite{Yu2018}, the authors utilize a 2D detector to seed 2D box annotations further refined by humans and report an increase of 60\% in the overall labeling speed. The authors in \cite{Acuna2018} train a recurrent CNN to predict polygons on an image to accelerate semantic segmentation tasks. A follow-up work \cite{Ling2019} further improves the system by predicting all polygon vertices simultaneously and facilitates real-time interaction. In \cite{Lee2018}, the authors propose a 3D labeling interface that enables the users selecting spatial seeds to infer segmentation, 3D centroid, orientation, and extent using pretrained networks. In \cite{Feng2019Active}, 2D labels are used to seed a LIDAR-based detector combined with human annotation based on uncertainty. All mentioned works are active learning frameworks in which a human is assisted by predictive models. Instead, we aim to investigate how well an \emph{automatic} pipeline with geometric verification can perform in this context.

Independently from our research, several recent works focused on differentiable rendering have been published. The works \cite{Loper2014, Kato2018, Chen2019} discuss different ways to produce gradients for rasterization of triangle-based meshes. The study described in \cite{Liu2019} represents a soft rasterizer approach in which each pixel is softly assigned to each triangle in the mesh, followed by a softmax to emulate z-buffering. In contrast, the authors in \cite{Li2018DMC} propose a path tracing approach towards differentiable rendering of triangle meshes. 

Concerning the more practically relevant research works, in \cite{Kundu2018, Manhardt2019} the authors employ a learned shape space from PCA or CAEs to predict the shape of cars. Nonetheless, their shape space has been either not differentiable in an end-to-end manner, or they use finite differences of local samples to approximate the gradient aiming to avoid a complex back-propagation through the rasterizer. Unlike these works, our differentiable rendering approach utilizes a shape space derived from DeepSDF \cite{park2019deepsdf}, which enables backpropagation onto a smooth shape manifold. Therefore, our method avoids the typical topology problems inherent to the related mesh-based approaches.

DensePose \cite{Guler2018} introduces a framework to densely map canonical 2D coordinates to human bodies. The subsequent works \cite{Kanazawa18, Zhang2019, Zuffi2019} discuss how to employ such coordinates with differentiable rendering in the wild to recover the pose, shape, texture and even future motion for different entities from localized crops. The study in \cite{Kulkarni2019csm} presents impressive results using a similar approach trained with a cycle consistency and differentiable rendering. These methods allow for projective scene analysis up to scale and cannot be immediately used for 3D automotive purposes. 

Dense coordinates have recently been extended to 3D space. The authors in \cite{zakharov2019dpod, park2019pix2pose, li2019cdpn, jafari2018ipose, peng2019pvnet} apply such representations for monocular pose estimation of known CAD models. In \cite{Wang2019NOCS}, the authors learn to predict normalized coordinates (NOCS) for object categories in RGB data and recover the metric scale and pose using an additional depth map via 3D-3D correspondences. 

In \cite{Engelmann2016}, the authors use stereo depth and run a detector to initialize the instances which are further optimized for pose and shape via SDF priors. In a follow-up work \cite{Engelmann2017}, the authors extend the framework with temporal priors to simultaneously recover smooth shapes and pose trajectories. Finally, \cite{Stutz2018} explores 3D object completion using a shape space and LIDAR as weak supervision with a probabilistic formulation. This work assumes correct localization and is focused solely on the reconstruction quality.
\section{Methodology}

We first discuss our shape space construction and the coupling into the CSS representation. Afterwards, we introduce our differentiable rendering approach tailored towards implicit surface representations. Eventually, our autolabeling pipeline is described in more detail. 


\begin{figure}[t]
	\centering
	\includegraphics[width=1\linewidth]{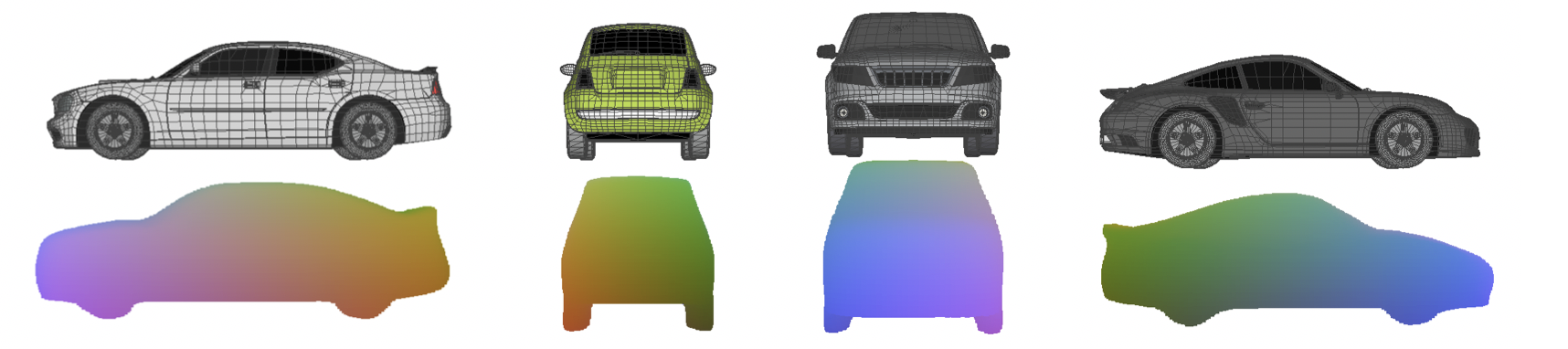}
	\caption{CSS representation. Top: Car models from the PD dataset \cite{paralleldomain}. Bottom: The same cars in the CSS representation: decoded shape vector $\mathbf{z}$ colored with NOCS.}
	\label{fig:car_shapes}  
\end{figure}

\subsection{Coordinate Shape Space}
We employ DeepSDF \cite{park2019deepsdf} to embed watertight car models into a joint and compact shape space representation within a single neural network. The idea is to transform input models into signed distance fields in which each value corresponds to a distance to the closest surface, with positive and negative values representing exterior/interior area. Eventually, DeepSDF forms a shape space of implicit surfaces with a decoder $f$ that can be queried at spatially-continuous 3D locations $\mathbf{x} = \{x_1, ..., x_N\}$ using the provided latent code $\mathbf{z}$ to retrieve SDF values $\mathbf{s} = \{s_1, ..., s_N\}$ as follows:
\begin{equation}
f(\mathbf{x};\mathbf{z}) = \mathbf{s}.    
\end{equation}

To facilitate approximate \textit{deformable shape matching}, we combine the shape space with NOCS \cite{Wang2019NOCS} to form the Coordinate Shape Space (CSS). To this end, we resize our models to unit diameter and interpret the 3D coordinates of the 0-level set as dense surface descriptions. 

To train $f$, we use a synthetic dataset provided by Parallel Domain \cite{paralleldomain}, which comprises car CAD models, as well as rendered traffic scenes with ground truth labels. Other synthetic datasets (\eg, CARLA \cite{Dosovitskiy17} \& VKITTI \cite{gaidon2016virtual}) could be used here as well. We trained on a subset of 11 models and with a latent code dimensionality of 3. We follow the original DeepSDF training, but project the latent vector onto the unit sphere after each iteration. In Figure \ref{fig:car_shapes}, we depict example models with their CSS representations.

\begin{figure}
	\centering
	\includegraphics[width=1\linewidth]{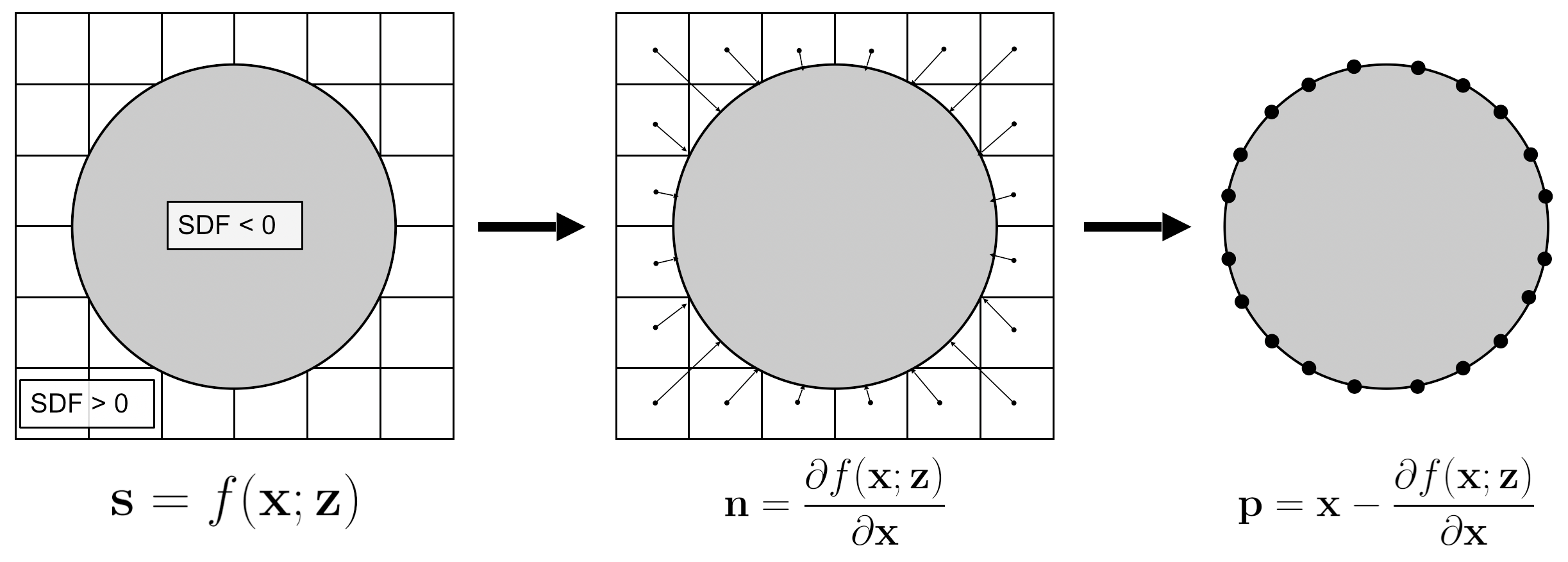}
	\caption{Surface projection. DeepSDF outputs the signed values $\mathbf{s}$ for query locations $\mathbf{x}$. Normals $\mathbf{n}$ can be computed analytically by a single backward pass. Given the signed values and normals, we project the query locations onto the object surface points $\mathbf{p}$. Only exterior points are visualized.}
	\label{fig:sdf_projection}  
\end{figure}

\subsection{Differentiable SDF Rendering}
An essential component of our autolabeling pipeline is the possibility to optimize objects with respect to the pose, scale, and shape. To this end, we propose, to the best of our knowledge, the first differentiable renderer for signed distance fields. Our renderer avoids mesh-related problems such as connectivity or intersection, but necessitates implementing a different approach for sampling the representation. Rendering implicit surfaces is done either with raytracing \cite{Curless1996} or variants of Marching Cubes \cite{Lorensen1987}. Here, we present an alternative that lends itself to backpropagation. 

\paragraph{Projection of 0-Isosurface}
Given query points $x_i$ and associated signed distance values $s_i$, we need a differentiable way to access the implicit surface encoded by $\mathbf{z}$. Simply selecting query points based on their distance values do not form a derivative with respect to the latent vector $\mathbf{z}$. Moreover, the regularly-sampled locations can be estimated only approximately on the surface. However, we utilize that deriving SDFs with respect to their location yields a normal at this point, practically computed in a backward pass:

\begin{equation}
    n_i = \frac{\partial f(x_i;\mathbf{z})}{\partial x_i}.
\end{equation}

As normals outline the direction to the closest surface and signed distance values provide the exact distance, we project the query location onto a 3D surface position $p_i$:

\begin{equation}
    p_i = x_i - \frac{\partial f(x_i;\mathbf{z})}{\partial x_i} f(x_i;\mathbf{z}).
\end{equation}

To obtain clean surface projections we disregard all points $x_i$ outside a narrow band ($|s_i| > 0.03$) of the surface. A schematic explanation can be found in Figure \ref{fig:sdf_projection}. With this formulation, we can define derivatives at $p_i$ with respect to the scale, pose, or latent code.


\paragraph{Surface Tangent Discs}
In the computer graphics domain, the concept of surface elements (surfels) \cite{Pfister2000} is a well-established alternative to connected triangular primitives. Our differentiable SDF representation yields oriented points and can be immediately used to render surface discs. To obtain a water-tight surface, we determine disk diameters large enough to close holes. In Figure \ref{fig:lm_examples} we outline the difference between (oriented) surface tangent discs and billboard ones pointing straight at the camera.

We construct the surface discs with the following steps:
\begin{enumerate}
    \item Given the normal of a projected point $n_i = \frac{\partial f(p_i; \mathbf{z})}{\partial p_i}$, we estimate the 3D coordinates of the resulting tangent plane visible in the screen. The distance $d$ of the plane to each 2D pixel $(u,v)$ can be computed by solving a system of linear equations for the plane and camera projection, resulting in the following solution (we refer to the supplement for details):
        \begin{equation}
        d = \frac{n_i \cdot p_i}{n_i \cdot K^{-1} \cdot (u,v,1)^T},
        \end{equation}
        where $K^{-1}$ is the inverse camera matrix, followed by backprojection to get the final 3D plane coordinate:
        \begin{equation}
            P = K^{-1} \cdot (u \cdot d, v \cdot d, d)^T.
        \end{equation}

    \item Estimate the distance between the plane vertex and surface point and clamp if it is larger than a disc diameter:
        \begin{equation}
        M = max(diam - || p_i - P||_2, 0)
        \end{equation}    
\end{enumerate}

To ensure water-tightness we compute the diameter from the query location density: $diam = \min_{i \neq j} ||x_i-x_j||_2 \sqrt{3}$. Executing the above steps for each pixel yields a depth map $\bm{D}_i$ and a tangential distance mask $\bm{M}_i$ at point $p_i$.

\begin{figure}
\centering
        \begin{subfigure}[b]{0.49\columnwidth}
        \centering
        \includegraphics[width=\linewidth]{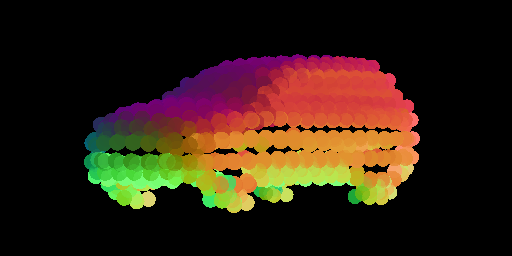}
        \end{subfigure}
        \begin{subfigure}[b]{0.49\columnwidth}
        \centering
        \includegraphics[width=\linewidth]{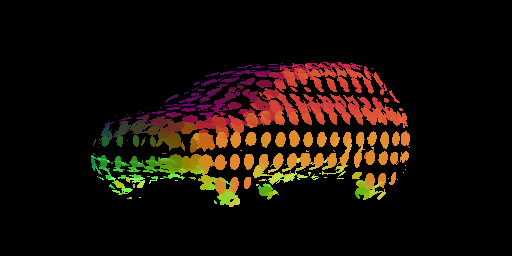}
        \end{subfigure}\hfill
        \caption{Oriented tangent discs (right) represent the surface geometry more accurately than billboard ones. We reduced spatial sampling and diameters for better emphasis.
        }\label{fig:lm_examples}
\end{figure}

\paragraph{Rendering Function}
To generate a final rendering we need a function to compose layers of 2D-projected discs onto the image plane. Similarly to \cite{Liu2019}, we combine colors from different point primitives based on their depth values. The closer the primitive is to the camera, the stronger its contribution. We use softmax to ensure that all primitive contributions sum up to 1 at each pixel. More specifically, the rendering function is:


\begin{equation}
    \mathcal{I} = \sum_i NOCS(p_i) * w_i ,
\end{equation}
where $\mathcal{I}$ is the resulting image, $NOCS$ returns coordinate coloring, and $w_i$ are the weighting masks that define the contribution of each disc:

\begin{equation}
    w_i =  \frac{exp(- \bm{\tilde{D}}_i \bm{\sigma} ) \bm{M}_{i}}{\sum_j {exp(- \bm{\tilde{D}}_j \bm{\sigma}) \bm{M}_{j}}},
\end{equation}
where $\bm{\tilde{D}}$ is the normalized depth, and $\bm{\sigma}$ is a transparency constant with $\bm{\sigma} \xrightarrow{} \infty$ being completely opaque as only the closest primitive is rendered. This formulation enables gradient flow from pixels to surface points and allows image-based optimization.

\begin{figure}[!t]
	\centering
	\includegraphics[width=1\linewidth]{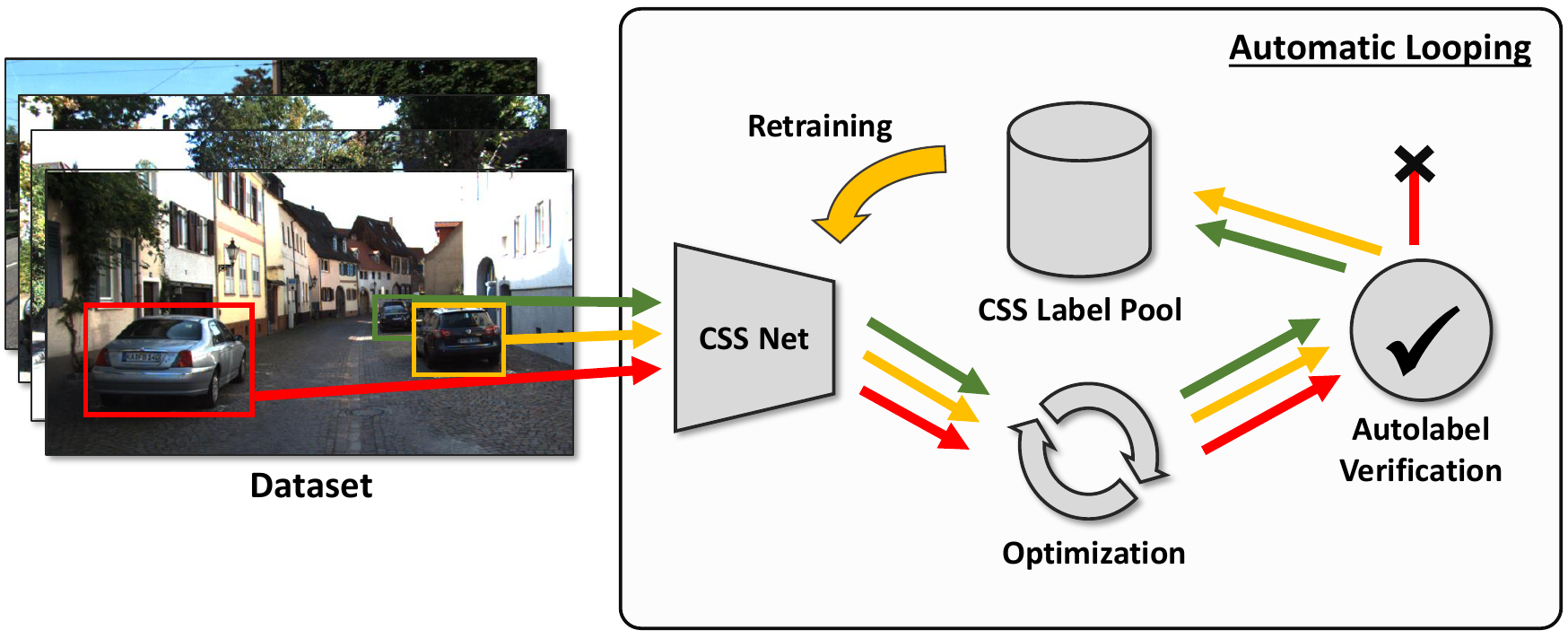}
	\caption{Automatic annotation pipeline. We fetch frames from the dataset and separately process each 2D detection using our CSS network and differentiable optimization procedure. Afterwards, we perform a verification to discard incorrect autolabels before saving them into our CSS label pool. Once all frames are processed, we retrain our CSS network and begin the next loop over the dataset.}
	\label{fig:pipeline}  
\end{figure}

\subsection{3D Autolabeling Pipeline}
The general idea of our autolabeling approach is to exploit weak labels and strong differentiable priors to recover labels of higher complexity. While this idea is generic, we focus specifically on cuboid autolabeling of driving scenes.

We present a schematic overview in Figure \ref{fig:pipeline} where we run multiple loops of the annotation pipeline. In the first loop, the CSS label pool solely consists of synthetic labels and the trained CSS network is therefore not well-adapted to real imagery. The results are noisy NOCS predictions which are reliable only for \textit{well-behaved} object instances in the scene. Therefore, we define a curriculum in which we first focus on \textit{easy} annotations and increase the difficulty over more loops. We define the difficulty of a label by measuring the pixel sizes, the amount of intersection with other 2D labels, and whether the label touches the border of an image (often indicating object truncation). We also establish thresholds for these criteria to define a curriculum of increasing difficulty.

\paragraph{CSS Network} The network is derived from ResNet18 and adopts an encoder-decoder structure, processing 128$\times$128 input patches to output a NOCS map of the same size and a 3D shape vector. Before executing the first annotation loop, our CSS network must learn to infer 2D NOCS maps and shape vectors from patches. As mentioned, we bootstrap such a mapping from a synthetic dataset. In total, we extract around 8k patches and, having access to CAD models, also create the necessary regression targets. We demonstrate some frames and training data in Figure \ref{fig:pd} and provide the additional information in the supplement.



\begin{figure}[!t]
	\centering
	\includegraphics[width=1\linewidth]{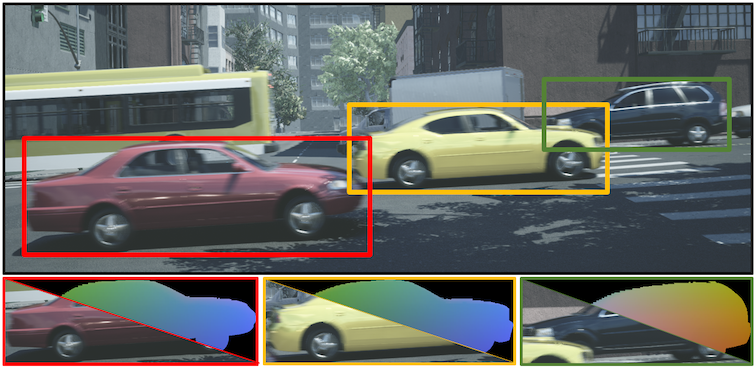}
	\caption{Synthetic PD dataset. Top: Random RGB frame. Bottom: Patches used for CSS training.}
	\label{fig:pd}  
\end{figure}

\subsubsection{Initialization and Optimization}
Here, we describe the process represented in Figure \ref{fig:intro} in more detail. For a given patch we infer the 2D NOCS map $\mathcal{M}$ and shape vector $\bm{z}$. We decode $\bm{z}$ into an SDF and retrieve the 3D surface points $\mathbf{p} = \{p_1, ..., p_n\}$ of the object model (as described in Section 3.2) in its local frame, for which we compute the NOCS coordinates $\mathbf{p}^c = \{p_1^c, ..., p_n^c\}$. We also project the 3D LIDAR points $\mathbf{l} = \{l_1, ..., l_k\} $ contained inside the camera frustum onto the patch and collect the corresponding NOCS coordinates $\mathbf{l}^c$. To estimate an initial pose and scale, we establish 3D-3D correspondences between $\mathbf{p}$ and $\mathbf{l}$. For each $p_i$, we find its nearest neighbor based on NOCS distances:
\begin{equation}
j^* = \argmin_j ||p_i^c - l_j^c||_2
\end{equation}
and keep it if $||p_i^c - l_{j^*}^c|| < 0.2$. Finally, we run Procrustes \cite{Schonemann1966} with RANSAC to estimate pose $(R,t)$ and scale $s$.

On this basis, we apply our differentiable optimization over complimentary 2D and 3D evidence. While the projective 2D information provides strong cues about the orientation and shape, 3D points allow reasoning over the scale and translation. At every iteration we decode the current shape vector estimate $\hat{\bm{z}}$, extract surface points $p_i$, and transform them based on the current estimates of the pose and scale:
\begin{equation}
\hat{p}_i = (\hat{R} \cdot \hat{s}) \cdot p_i  + \hat{t}.  
\end{equation}

Given these surface model points in the scene frame, we compute the individual losses as follows.

\paragraph{2D loss:} We employ our differentiable SDF renderer to produce a rendering $\mathcal{R}$ for which we seek maximum alignment with NOCS map $\mathcal{M}$. Since our predicted $\mathcal{M}$ can be noisy (especially in the first loop), minimizing dissimilarity $min ||\mathcal{M} - \mathcal{R}||$ can yield suboptimal solutions. Instead, for each rendered spatial pixel $r_i$ in $\mathcal{R}$ we determine the closest NOCS space neighbor in $\mathcal{M}$, named $m_j$ within a small radius $\theta$, and set them in correspondence if their NOCS distance is below a predefined threshold. 
The loss is then defined as the mean distance over all such correspondences $C_{2D}$ in the NOCS space:
\begin{equation}
loss_{2D} = \frac{1}{|C_{2D}|} \sum_{(i,j) \in C_{2D}} || \mathcal{R}(r_i) -  \mathcal{M}(m_i)||_2.
\end{equation}

\paragraph{3D loss:} For each $\hat{p}_i$, we determine the nearest neighbor from $l$ and keep it if it is closer than $0.25$m. As the initializations are usually good, we avoid outliers in the optimization with such a tight threshold. The loss is then calculated as the mean distance over all correspondences $C_{3D}$: 

\begin{equation}
loss_{3D} =  \frac{1}{|C_{3D}|} \sum_{(i,j) \in C_{3D}} || \hat{p}_i - l_j ||_2.
\end{equation}
\newline
Altogether, the final criterion is the sum of both losses:
\begin{equation}
loss = loss_{2D} + loss_{3D}.
\end{equation}
As the loss terms have similar magnitudes we did not consider a need for any additional balancing. 

\subsubsection{Verification and CSS Retraining}
Our optimization framework will inevitably output incorrect results at times, so we need to ensure that the influence of badly-inferred autolabels is minimized. To this end, we enforce geometric and projective verification aiming to remove incorrect autolabels with the largest impact. To achieve this, we measure the amount of LIDAR points that are in a narrow band ($0.2$m) around the surface of an autolabel and reject it if less than 60\% are outside this band. Furthermore, we define a projective constraint where autolabels should be rejected if the rendered mask's IoU with the provided 2D label is below $70\%$.

All autolabels that remain after the verification stage are gathered and added to the CSS label pool. After the first loop, we obtain a mixture of synthetic and real samples that are then used to retrain and improve the robustness of our CSS network. Over multiple self-improving loops, we keep growing and retraining, which results in better CSS predictions, better initializations, and more accurate autolabels.

\begin{figure*}[t]
	\centering
	\includegraphics[width=1\linewidth, height=8cm]{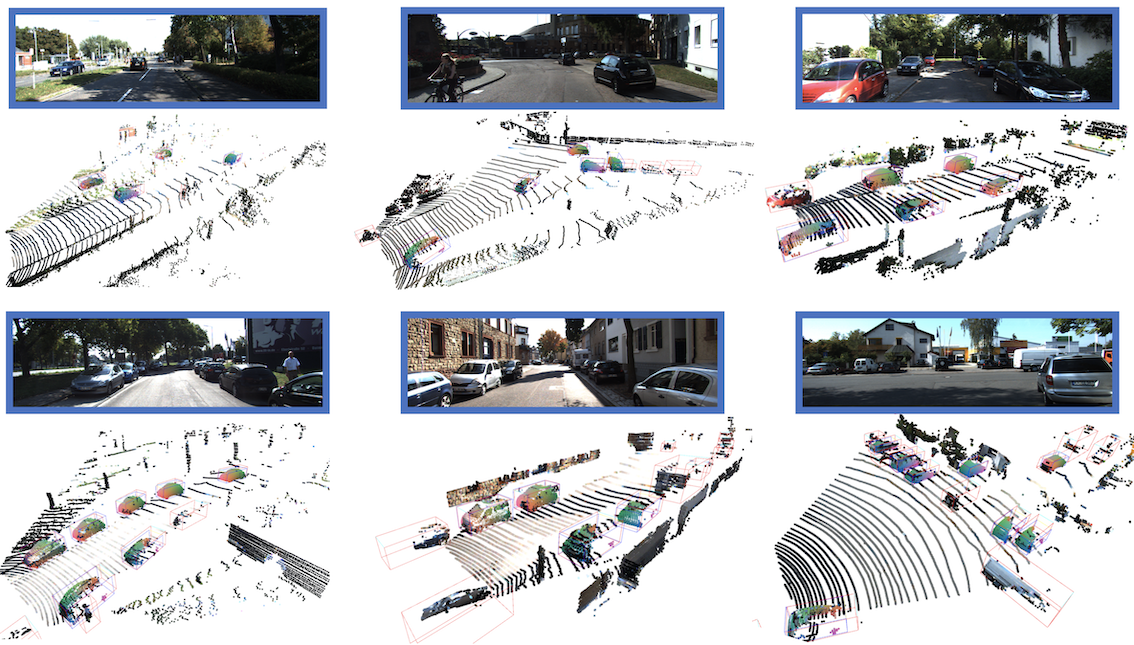}
	\caption{Qualitative results of our labeling pipeline. We mark the ground truth cuboids in red and our predictions in blue. We can achieve rather tight fits that lead to cuboids of slightly different sizes compared with the ground truth.}
	\label{fig:eval_autolabels}  
\end{figure*}

\section{Experimental Evaluation}

We evaluate our approach on the well-established KITTI3D dataset \cite{Geiger2012}, including 7481 frames with accompanied cuboid labels for the ``Car'' category, which we focus on.
We consider the most-widely used 3D metrics for driving datasets: BEV IoU and 3D IoU from KITTI3D as well as the distance-thresholded metric (NS) from NuScenes~\cite{nuscenes2019}, which decouples location from scale and orientation. All three metrics are utilized to evaluate Average Precision for matches at certain cutoffs, and we threshold BEV and 3D IoUs at 0.5 whereas NS is computed at distance cutoffs of 0.5m and 1m. 

The KITTI3D metrics are often evaluated at a strict threshold of 0.7. After thorough inspection, we observed that it is difficult to infer the correct cuboid size from tight surface fits. The KITTI3D cuboids have a varying amount of spatial padding and 3D detection methods learn these offsets. Therefore, we opted to relax the threshold aiming to facilitate a fairer comparison with respect to the estimated tight cuboids. Figure~\ref{fig:eval_autolabels} represents several examples of appropriate autolabel estimates among which some do not pass the 0.7 3D IoU criterion. 

\setlength{\tabcolsep}{4pt}
\begin{table*}[!t]
  \centering
    \resizebox{1\linewidth}{!}{%
    \begin{tabular}{c|c|cccc|cccc|cccc}
    \toprule
    \multirow{2}[4]{*}{\textbf{Loop}} & \multirow{2}[4]{*}{\textbf{Diff.}} & \multicolumn{4}{c|}{\textbf{KITTI GT}} & \multicolumn{4}{c|}{\textbf{RCNN}} & \multicolumn{4}{c}{\textbf{MASK-RCNN}} \\
\cmidrule{3-14}          &       & \textbf{BEV@0.5} & \textbf{3D@0.5} & \textbf{NS@0.5} & \textbf{NS@1.0} & \textbf{BEV@0.5} & \textbf{3D@0.5} & \textbf{NS@0.5} & \textbf{NS@1.0} & \textbf{BEV@0.5} & \textbf{3D@0.5} & \textbf{NS@0.5} & \textbf{NS@1.0} \\
    \midrule
1     & E     & 78.09 & 63.53 & 85.59 & 95.58 & 78.45 & 63.71 & 85.85 & 95.62 & 78.46 & 63.69 & 86.27 & 95.76 \\
    \midrule
    \multirow{2}[2]{*}{2}   & E     & 77.84 & 62.25 & 82.40 & 90.84 & 80.57 & 60.11 & 86.05 & 94.62 & 80.70 & 63.96 & 86.52 & 94.31  \\
                            & M     & 59.75 & 42.23 & 60.27 & 77.91 & 61.17 & 42.37 & 64.11 & 85.85 & 63.36 & 44.79 & 64.44 & 85.24 \\
    \bottomrule
    \end{tabular}%
    }
    \caption{Cuboid autolabel quality when inputting into the CSS network (a) 2D ground truth boxes, (b) RCNN detections, and (c) Mask-RCNN detections. We run two self-improving loops to slowly incorporate more labels into the pool.}
  \label{tab:full_pipeline}%
\end{table*}%

\paragraph{Implementation Details} We use PyTorch~\cite{paszke2017automatic} to implement the whole pipeline. For each 2D-labeled instance we run 50 iterations and use the ADAM optimizer for the pose variables with a learning rate of $0.03$ whereas SGD is applied to the scale and shape with smaller learning rates ($0.01$ and $0.0005$) and no momentum to avoid observed overshooting behavior. It takes approximately 6 seconds to autolabel a single instance on a Titan V GPU, and one autolabeling loop takes 1-2 hours to complete for all frames when parallelizing on 2 GPUs. 

\paragraph{Data Augmentation} Our synthetic bootstrapping requires many kinds of augmentation to allow for initial domain transfer. We utilize an extensive set of functions, including random rotations up to 10$^{\circ}$, horizontal flips, and random cropping. This allows not only to considerably expand the data size, but also enables explicitly covering cases of partial occlusion and truncation. Additionally, we vary the brightness, contrast, and saturation of input patches. Moreover, as we have access to corresponding CAD models, we render surface normals and use Phong shading to generate geometrically-sensible relighting.

\subsection{Correctness of autolabels}
The most important quantitative criterion is the actual correctness of the estimated cuboids. Although our method is fully automatic, we have access to KITTI3D 2D ground truth boxes and therefore evaluate two scenarios. Firstly, we obtain 2D boxes from KITTI3D for autolabeling and then, utilize their predefined criteria to determine whether an annotation is considered easy or moderate. Secondly, we employ the detectron2 implementation \cite{wu2019detectron2} of Mask-RCNN \cite{He2017} using a ResNeXt101 backbone trained on COCO to evaluate the applicability of off-the-shelf object detectors for full automation. In the detection scenario, we run separate experiments for boxes and masks and apply following difficulty criteria similar to KITTI3D: easy if label height $> 40$px and not touching other labels or image borders; moderate if height $> 25$px and not having an IoU $> 0.30$ with any other label.



\begin{figure}[b]
	\centering
	\includegraphics[width=1\linewidth]{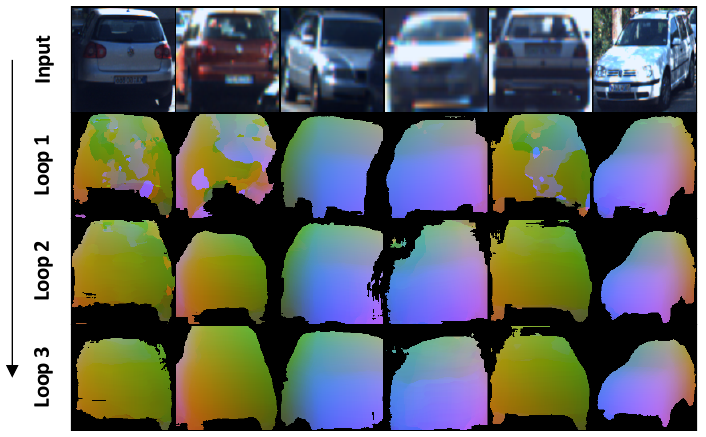}
	\caption{NOCS prediction quality of our network over consecutive loops for the same patch. Initially, the predictions are rather noisy because of the synthetic domain gap. Within each subsequent autolabeling loop the predictions become more accurate overall.}
	\label{fig:eval_nocs}  
\end{figure}

We present the obtained results in Table \ref{tab:full_pipeline}. As expected, the first loop with a purely synthetically-trained inference on easy real samples does not yield a considerable difference between the three scenarios. All of them are impacted by noisy CSS predictions and start from the same RANSAC initialization, although the detection-derived labels are tighter and slightly less influenced by CSS background noise. Overall, each scenario achieves a \textbf{BEV AP close to 80\%} and a 3D AP of approximately 60\% over all easy samples. We execute a second loop over the dataset using the retrained CSS network and observe that the results for the easy samples stabilized. Additionally, we note that we can recover approximately 60\% BEV AP over all considered moderate samples. Noteworthy, the estimated NS scores are quite high, indicating that most autolabels (more than 90\%) are within one meter of the real location.

However, we observe a drop of around 20 points across all metrics for the harder samples. As our method is reliant on 3D-3D RANSAC, it requires a minimum set of inliers to achieve proper solutions. We often observe that this made correspondence finding difficult for occasional samples due to occlusion and distance, thus impacting recall. 

Figure \ref{fig:eval_nocs} shows the increasing quality of the estimated NOCS predictions over multiple loops. Overall, we observe a rather fast diffusion into the target domain and that executing two loops is sufficient to stabilize the results.

\setlength{\tabcolsep}{11pt}
\begin{table*}[t]
\centering
    \resizebox{1\linewidth}{!}{%
\begin{tabular}{r|cc|cc|cc}
\toprule
\textbf{} & \multicolumn{2}{c|}{\textbf{2D AP @ 0.5/0.7}} & \multicolumn{2}{c|}{\textbf{3D AP @ 0.5/0.7}} & \multicolumn{2}{c}{\textbf{BEV AP @ 0.5/0.7}} \\
\textbf{Method} & Easy & Moderate & Easy & Moderate & Easy & Moderate \\
\midrule

PointPillars \cite{Lang2019} (Original Labels) & - / - & - / - & 94.8 / 81.1 & 92.4 / 68.2 & 95.1 / 92.1 & 95.1 / 84.7 \\
PointPillars \cite{Lang2019} (Autolabels) & - / - & - / - & 90.7 / 22.4 & 71.1 / 13.3 & 94.9 / 81.0 & 88.5 / 59.8 \\
\midrule
MonoDIS \cite{Simonelli2019} (Original Labels) & 96.1 / 95.5 & 92.6 / 86.5 & 45.7 / 11.0 & 32.9 / 7.1 & 52.4 / 17.7 & 37.2 / 11.9 \\
MonoDIS \cite{Simonelli2019} (Autolabels) & 96.7 / 85.8 & 86.2 / 67.6 & 32.9 / 1.23 & 22.1 / 0.54 & 51.1 / 15.7 & 34.5 / 10.52\\
\bottomrule

\end{tabular}
}

\caption{The performance comparison of the 3D object detectors trained on the true KITTI labels vs. our autolabels. Concerning the BEV metric, \textbf{the detectors trained on autolabels alone achieve the results equal to the current state of the art}. In the case of the 3D AP metric, the competitive results are achieved in both considered variants at the IoU 0.5 threshold.}
\label{tab:detection}
\end{table*}

\subsection{Ablation}
We aim to investigate how much the initial estimates from 3D-3D RANSAC benefit from our optimization. To this end, we consider the easy ground truth boxes from KITTI3D and utilize the synthetic CSS network to analyze the first annotation loop with the worst initialization. As represented in Table \ref{tab:ablation_optim}, the RANSAC baseline provides rather good localization which is best captured by the NS metrics (81.36\% and 95.45\%). Nonetheless, the pose-optimized autolabels yield a significant jump in 3D IoU (41.85\% vs. 63.42\%), suggesting that we recover substantially better rotations, given that the NS scores are similar. When ablating over the other variables we observe rather mixed results in which certain metrics increase or decrease.

When ablating over the losses, we note a drastic drop in the 3D metrics when optimizing only in 2D. Intuitively, our differentiable renderer aligns the data rather appropriately in the image space; however, both scale and translation are freely drifting. Therefore, optimizing the 3D loss results in strong spatial alignment. Nonetheless, optimizing the sum of both losses trades BEV AP (80.61 to 78.09) for 3D AP (60.92 to 63.53).  

\begin{figure}[b]
	\centering
	\includegraphics[width=1\linewidth, trim=0 40mm  0mm 60mm, clip]{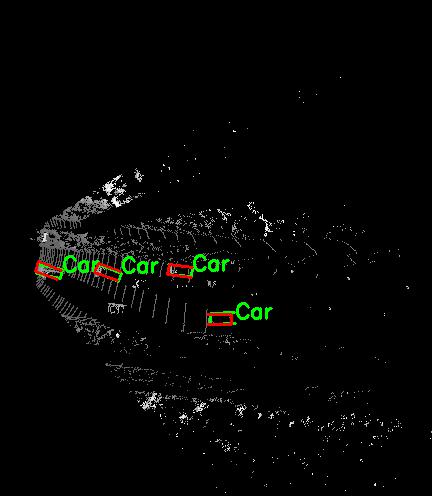}
	\includegraphics[width=1\linewidth]{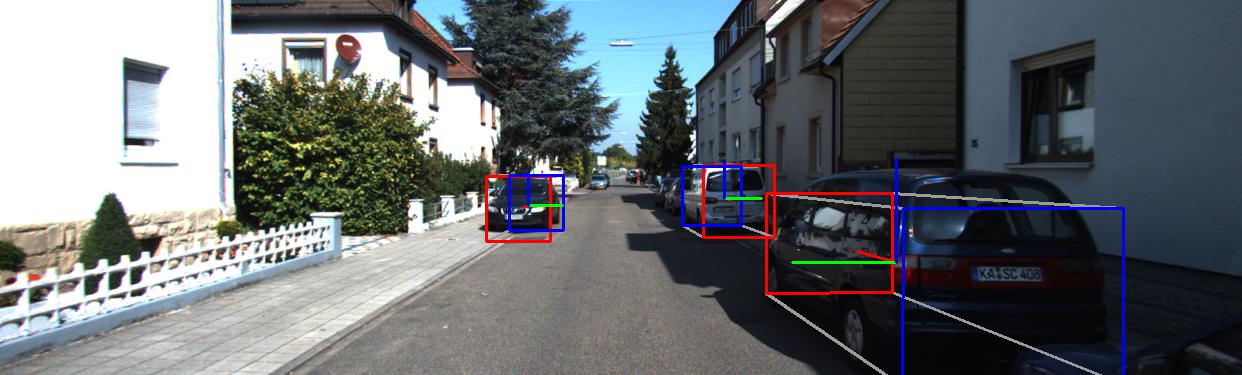}
	\includegraphics[width=1\linewidth]{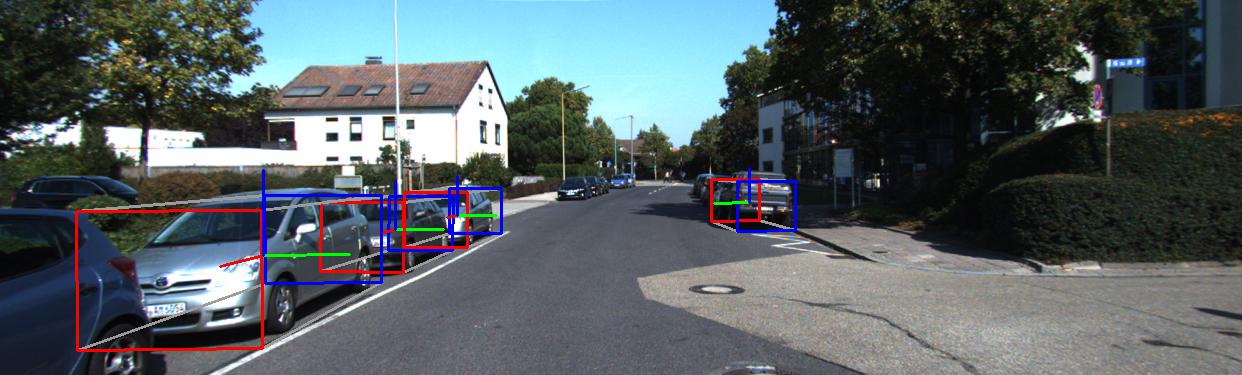}
	\caption{Detections from the autolabel-trained detectors. We draw local 3D frames to identify correct orientation.}
	\label{fig:eval_monodis}  
\end{figure}

\subsection{3D Object Detection}
Since autolabels are usually not the final goal but rather a means to an end, we investigate the applicability of our labels to the task of 3D object detection. We evaluate the quality of our labels concerning both a traditional LIDAR-based detection setting and a purely monocular setting, based on several recent works that achieved high quality results on the KITTI dataset \cite{Manhardt2019, Simonelli2019, Lang2019}. 

We implemented a version of the current state-of-the-art monocular detector MonoDIS \cite{Simonelli2019} and ensured that we can reproduce the reported results. Additionally, we utilize the official implementation of PointPillars \cite{Lang2019}, a state-of-the-art LIDAR-only detector. To train MonoDIS, we follow the training schedule proposed in \cite{Simonelli2019}. Concerning PointPillars training, we accelerate the training by means of 8 V100 GPUs and a batch size of 16. Accordingly, we scale the learning rate by a factor of 8. We evaluate the obtained results on the MV3D train/val split \cite{Chen2017}. While training on autolabels, we do not change any hyperparameters defined in the baseline protocols.

\setlength{\tabcolsep}{8pt}
\begin{table}[t]
  \centering
    \resizebox{1\linewidth}{!}{%
    \begin{tabular}{r|cccc}
    \toprule
    \textbf{Config} & \textbf{BEV@0.5} & \textbf{3D@0.5} &
    \textbf{NS@0.5} & \textbf{NS@1.0} \\
    \midrule
    RANSAC              & 77.00 & 41.85 & 81.36 & 95.45 \\
    \midrule
    $(R, t)$            & 77.19 & 63.42 & 86.20 & 95.53 \\
    $(R, t), s$         & 77.23 & 62.92 & 86.01 & 95.32 \\
    $(R, t), s, \bm{z}$ & 78.09 & 63.53 & 85.59 & 95.58 \\
    \midrule
    2D loss             & 18.08 & 11.09 & 18.35 & 46.19 \\ 
    3D loss             & 80.61 & 60.92 & 85.63 & 95.49 \\
    \bottomrule
    \end{tabular}%
    }
    \caption{Ablation study over each optimization variable and each separate loss.}
  \label{tab:ablation_optim}%
\end{table}%

We present the comparison results in Table \ref{tab:detection} and depict several qualitative detections obtained from autolabel-trained detectors in Figure \ref{fig:eval_monodis}.


Remarkably, concerning the BEV metric, both detectors trained \textit{on autolabels alone} achieve competitive performance compared with detectors trained on true KITTI labels at both the 0.5 and 0.7 IoU thresholds. This indicates that our autolabeling pipeline is capable of highly accurate localization of cuboids. 

Considering the 3D AP metric, the obtained results are in line with the conclusions represented in Table \ref{tab:full_pipeline}. At the more tolerant IoU 0.5 threshold, our autolabel trained detectors performed within $70 - 90\%$ of the true labels. Occasionally missing detections and poor shape estimates do not deteriorate the overall performance. 

At the IoU 0.7 threshold, the detector performance worsens. We observe that this is not caused by poor predictions, but by the fact that KITTI labels are often inflated with respect to the estimated cuboids. Therefore, at the strictest thresholded 3D IoU, we observe a corresponding drop in precision for detectors trained on our ``tight" autolabels.


\section{Conclusion}
We present a novel view on parametric 3D instance recovery in the wild based on a self-improving autolabeling pipeline, purely bootstrapped from synthetic data and off-the-shelf detectors. Fundamental to our approach is the combination of dense surface coordinates with a shape space, and our contribution towards differentiable rendering of SDFs. We show that our approach can recover a substantial amount of cuboid labels with high precision, and that these labels can be used to train 3D object detectors with results close to the state of the art.
Future work will be focused on investigating additional categories for parametric reconstruction, such as pedestrians or road surfaces.

{\small
\bibliographystyle{ieee_fullname}
\bibliography{egbib}
}

\clearpage
\newpage\appendix
\begin{center}
{\Large \textbf{Supplementary Material}}
\end{center}

\section{DeepSDF}
\begin{figure}[b]
	\centering
	\includegraphics[width=1\linewidth]{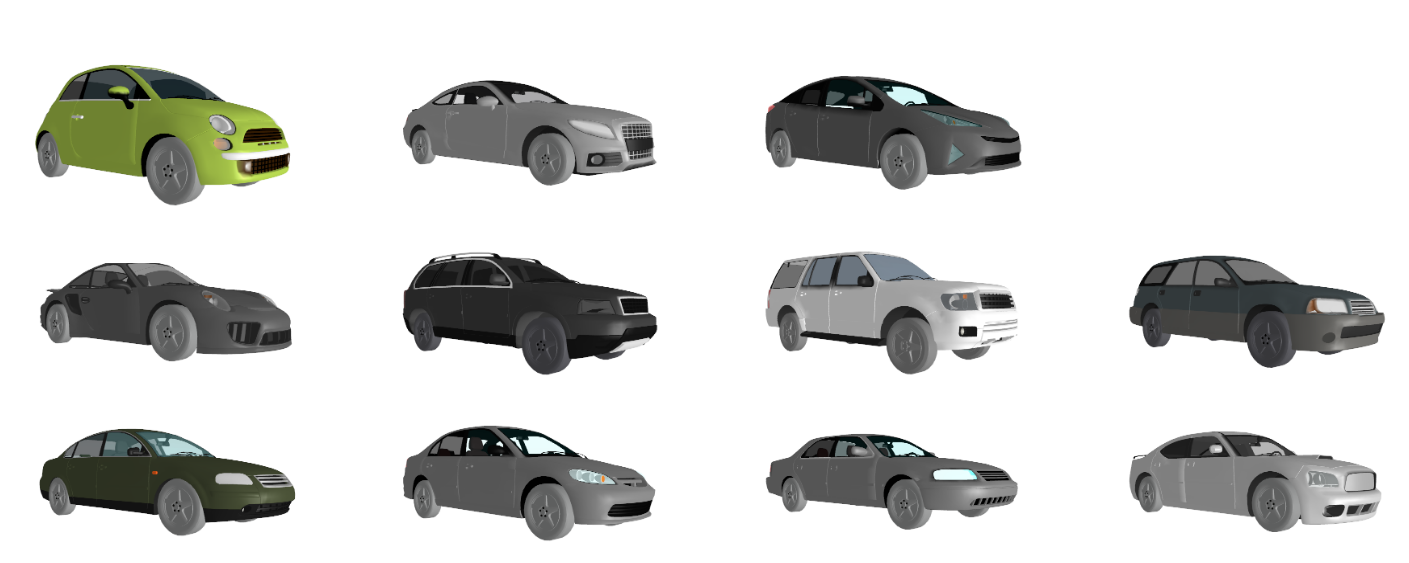}
	\caption{Cars from the PD dataset that were used to train our DeepSDF shape space.}
	\label{fig:pd_cars}  
\end{figure}

\begin{figure}[t]
	\centering
	\includegraphics[width=1\linewidth]{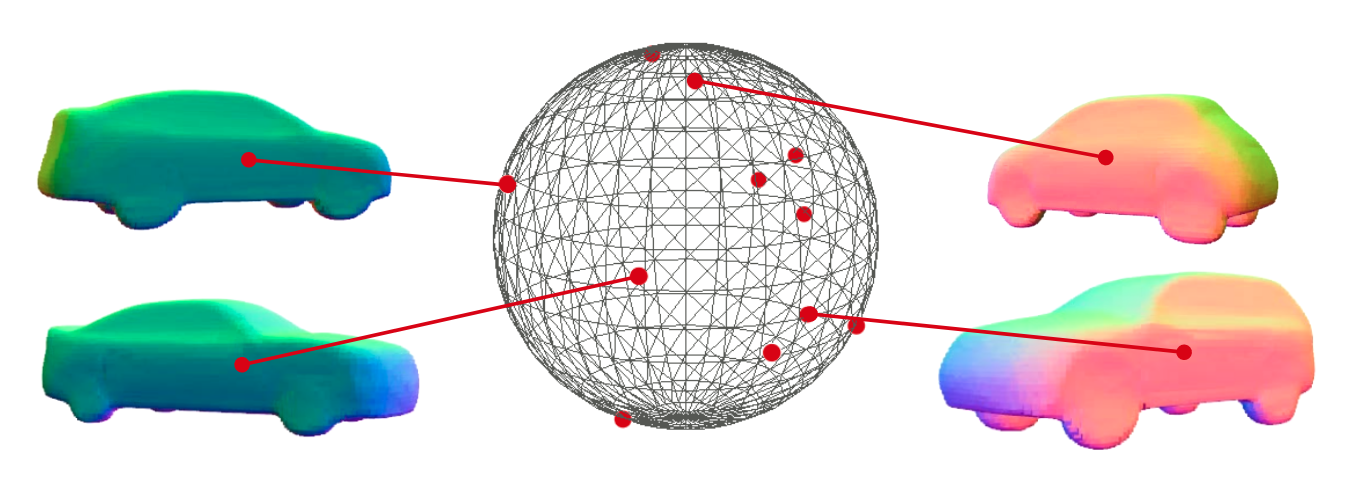}
	\caption{Visualization of the 3D latent vector space and the corresponding shapes after 0-isosurface projection.}
	\label{fig:sphere}  
\end{figure}

Our DeepSDF network is trained to cover a set of normalized predefined cars coming from the Parallel Domain (PD) dataset (visualized in Figure \ref{fig:pd_cars}). Since we trained our latent shape vectors $\mathbf{z}$ to be 3-dimensional, we can easily visualize their positions on the surface of a 3D sphere, as shown in Figure~\ref{fig:sphere}. Each red point represents a specific model shape embedded in the space.

\section{Data Generation}
\begin{figure}[b]
\centering
        \begin{subfigure}{0.33\columnwidth}
        \centering
        \includegraphics[width=\linewidth]{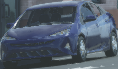}
        \caption{RGB}
        \label{fig:vis_rgb}
        \end{subfigure}\hfill
        \begin{subfigure}{0.33\columnwidth}
        \centering
        \includegraphics[width=\linewidth]{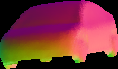}
        \caption{Normals}
        \label{fig:vis_normals}
        \end{subfigure}\hfill
        \begin{subfigure}{0.33\columnwidth}
        \centering
        \includegraphics[width=\linewidth]{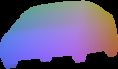}
        \caption{NOCS}
        \label{fig:vis_nocs}
        \end{subfigure}\hfill
        
        \begin{subfigure}{0.33\columnwidth}
        \centering
        \includegraphics[width=\linewidth]{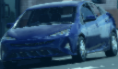}
        \caption{Light 1}
        \label{fig:vis_light1}
        \end{subfigure}\hfill
        \begin{subfigure}{0.33\columnwidth}
        \centering
        \includegraphics[width=\linewidth]{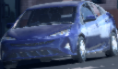}
        \caption{Light 2}
        \label{fig:vis_light2}
        \end{subfigure}\hfill
        \begin{subfigure}{0.33\columnwidth}
        \centering
        \includegraphics[width=\linewidth]{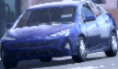}
        \caption{Light 3}
        \label{fig:vis_light3}
        \end{subfigure}\hfill
        \caption{Data input modalities: (a) input RGB image, (b) rendered normal map, (c) rendered NOCS. Light module outputs: (d, e, f). 
        }\label{fig:lm_examples}
\end{figure}

\begin{figure*}[!h]
	\centering
	\includegraphics[width=0.49\linewidth]{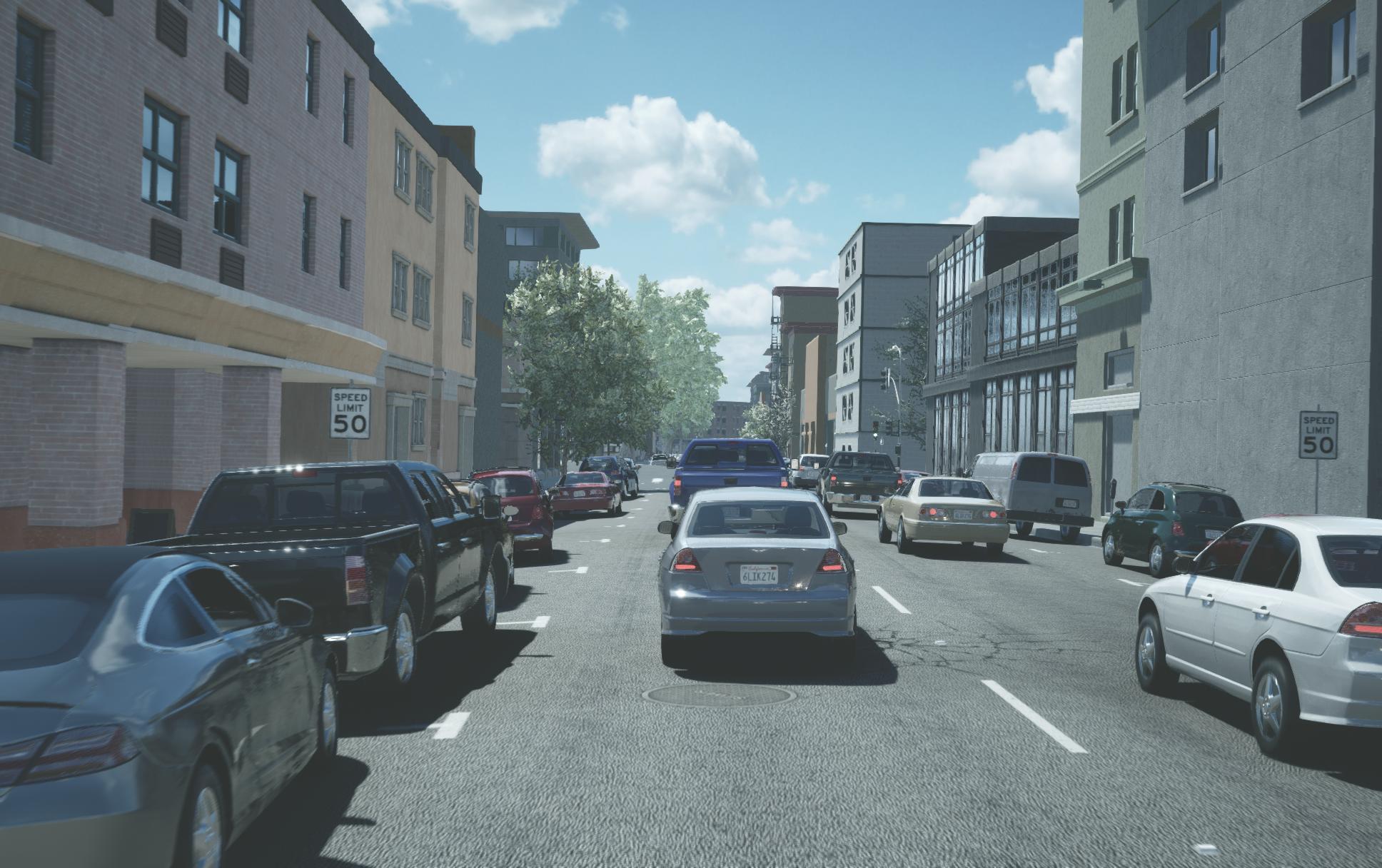}
	\includegraphics[width=0.49\linewidth]{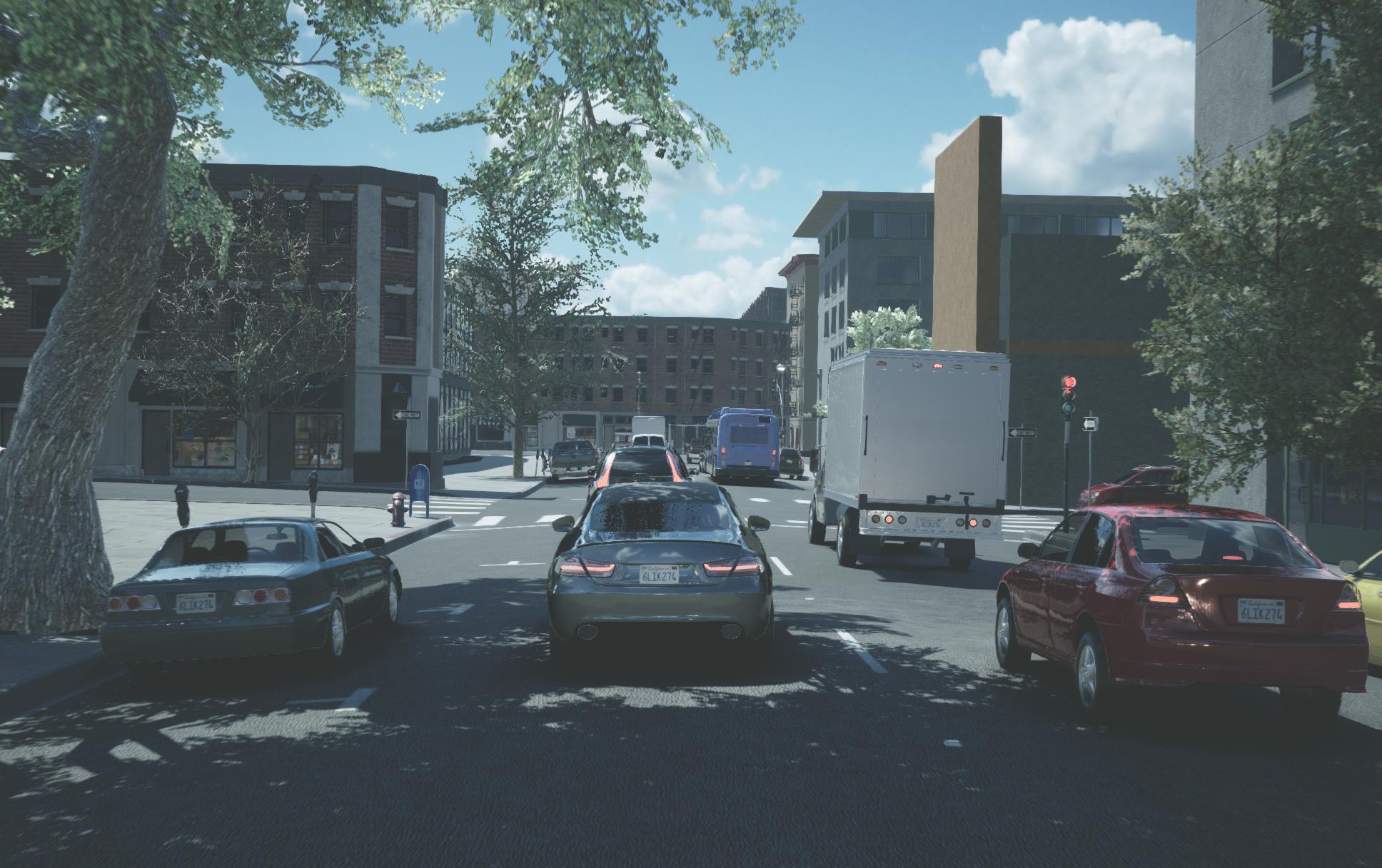}
	\includegraphics[width=0.49\linewidth]{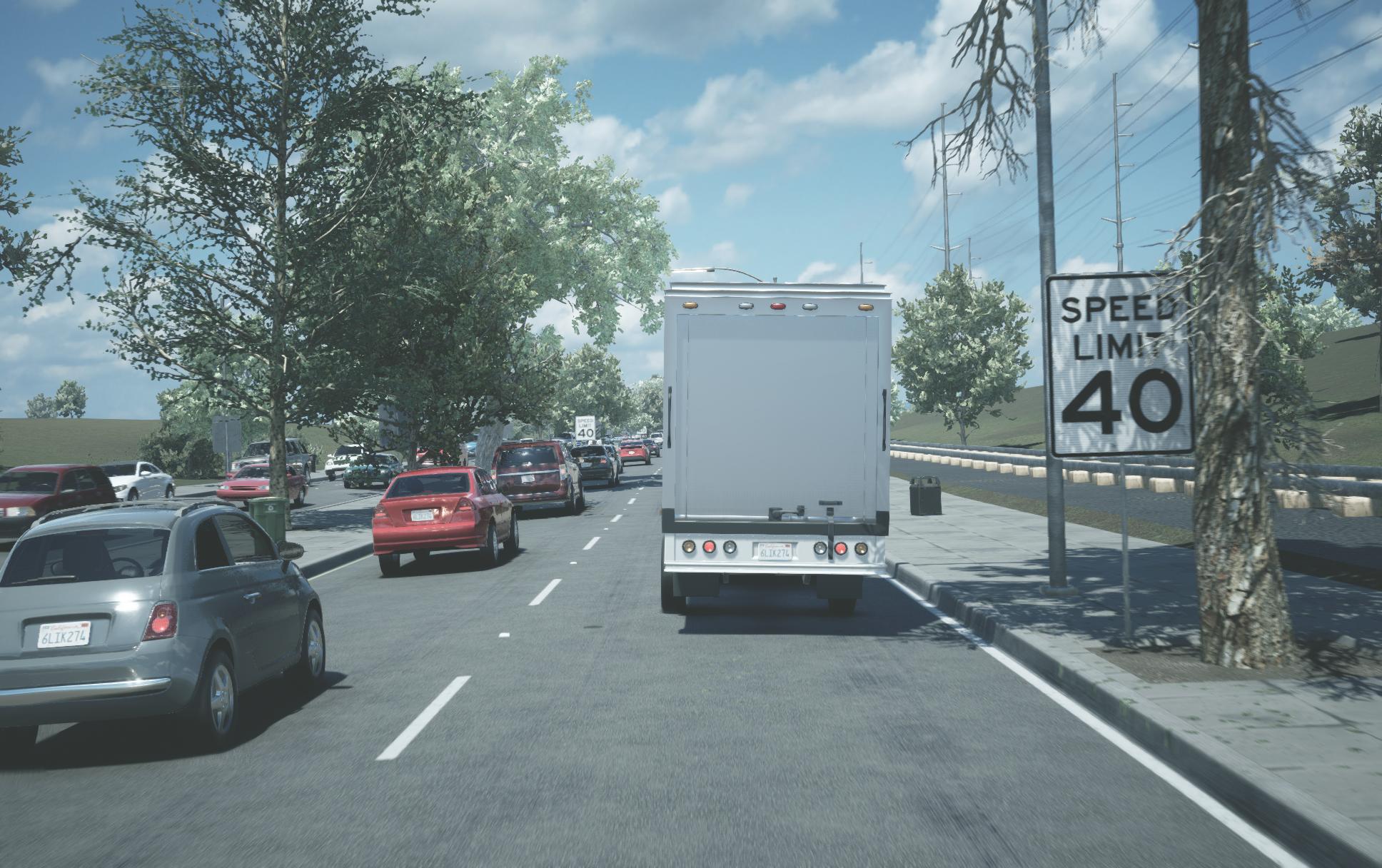}
	\includegraphics[width=0.49\linewidth]{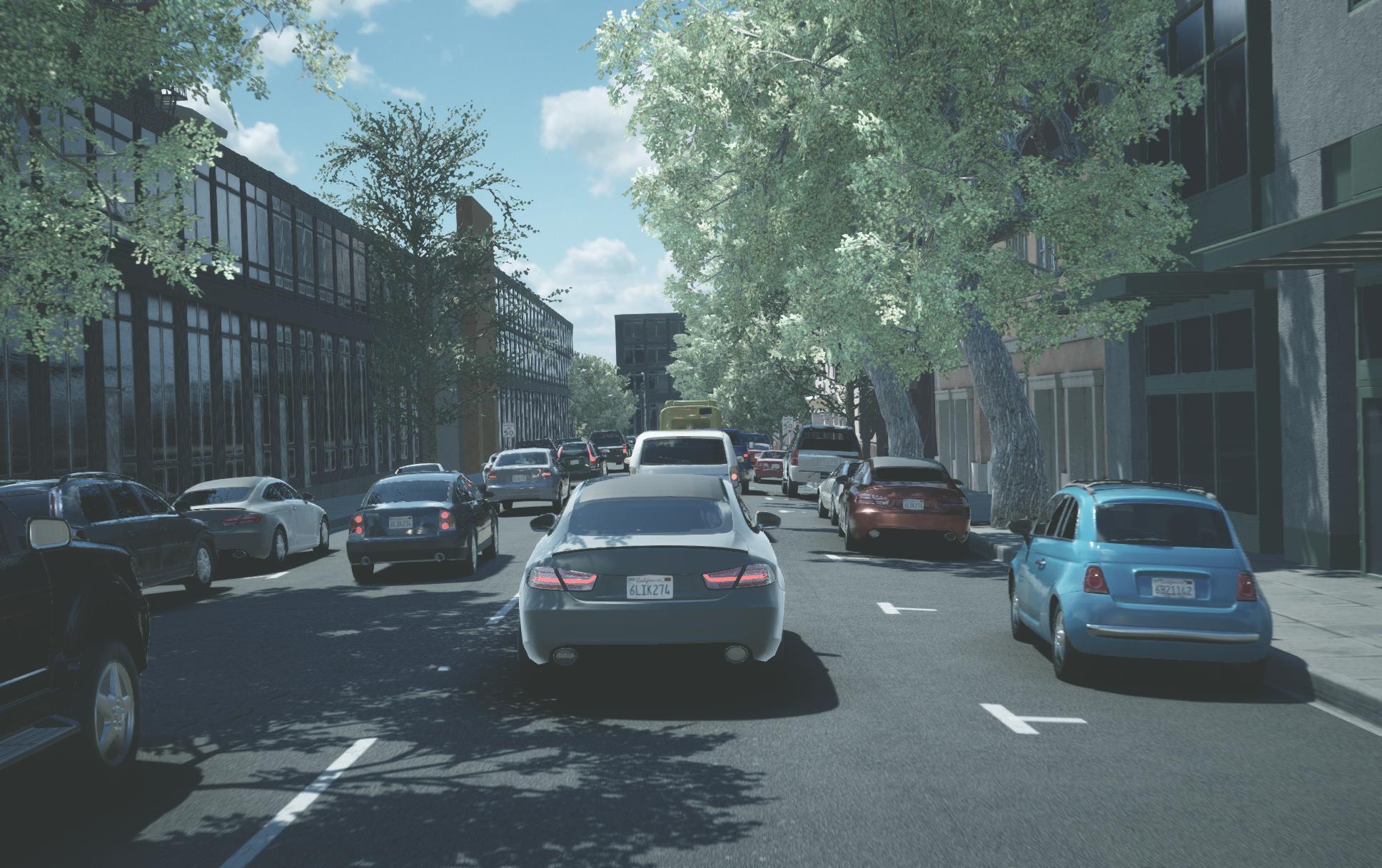}
	\caption{Some example images from the PD dataset that we used for synthetic CSS training.}
	\label{fig:pd_images}  
\end{figure*}

Given a set of ground truth poses with associated CAD models that we extract from our synthetic PD dataset (see Figure \ref{fig:pd_images}), we use our DeepSDF network and our differentiable renderer to project the models onto the screen. Instead of rendering the colors, we render the models' normalized coordinates (NOCS) represented as RGB channels (Figure~\ref{fig:vis_nocs}). Additionally, we render object normals (Figure~\ref{fig:vis_normals}), which are subsequently used for 0-isosurface projection. Our augmentation module takes RGB crops (Figure~\ref{fig:vis_rgb}) and normal maps and applies 2D and 3D augmentations. 2D augmentations are based on the \texttt{torchvision.transforms} module operations and include random 10$^\circ$ rotations, horizontal flips, cropping, changes in brightness, contrast, and saturation. Moreover, normal maps provide us with local surface information that we use in conjunction with simple Phong shading. Thus, we can generate lighting based on different illumination types (namely ambient, diffusive, and specular) during training. Examples are shown in Figures~\ref{fig:vis_light1}, \ref{fig:vis_light2}, \ref{fig:vis_light3}. Note how the bottom and top sides of the car change illumination between frames (e) and (f).

\section{Pipeline Components}

The comprehensive representation of our pipeline is depicted in Figure \ref{fig:pipeline}. Below, we describe its main components in detail.

\subsection{CSS Net}
The architecture of our CSS net is visualized in Figure~\ref{fig:pipeline} and we use a ResNet18 backbone architecture. The decoders use bilinear interpolation as an upsampling operation rather than deconvolution to decrease the number of parameters and the required amount of computations. Each upsampling is followed by concatenation of the output feature map with the feature map from the previous level, and one convolutional layer. Since the CSS net is trained on synthetic data, it is initialized with ImageNet weights and the first five layers are frozen in order to prevent overfitting to peculiarities of the rendered data. Five heads of the CSS net are responsible for the output of U, V, and W channels of the NOCS as well as the object's mask and its latent vector, encoding its DeepSDF shape.

\subsection{Pose Estimation Block}
The pose estimation block (see initial conditions estimation in (Figure~\ref{fig:pipeline}) is based on 3D-3D correspondence estimation. The procedure is defined as follows: Our CSS net outputs normalized object coordinates (NOCS), mapping each RGB pixel to a 3D location on the object's surface. NOCS are backprojected onto the LIDAR frustum points using the provided camera parameters. Additionally, our network outputs a latent vector, which is then fed to the DeepSDF net and transformed to a surface point cloud using the 0-isosurface projection. Since our DeepSDF is trained to output normalized models placed at the origin, each point on the resulting model surface represents NOCS. At this point, we are ready to proceed with pose estimation.

NOCS are used to establish correspondences between frustum points and model points. Backprojected frustum NOCS are compared to the predicted model coordinates, and nearest neighbors for each frustum point are estimated. RANSAC is used for robust outlier rejection. At each iteration we take 4 random points ($n$) from the set of the correspondences and and feed them to the Procrustes algorithm, giving us initial estimates for the pose and scale. 

The following RANSAC parameters are used: the number of iterations $k$ is based on a standard function of the desired probability of success $p$ using a theoretical result:
\begin{equation}
    k = \frac{\log(1 - p)}{\log(1 - w^n)},
\end{equation}
where $w$ is the inlier probability and $n$ are the independently selected data points. In our case $p = 0.9$ and $w = 0.7$.

We use a threshold of 0.2m to estimate the inliers and choose the best fit. The final pose and scale are computed based on the inliers of the best fit.

\begin{figure*}[t]
	\centering
	\includegraphics[width=1\linewidth]{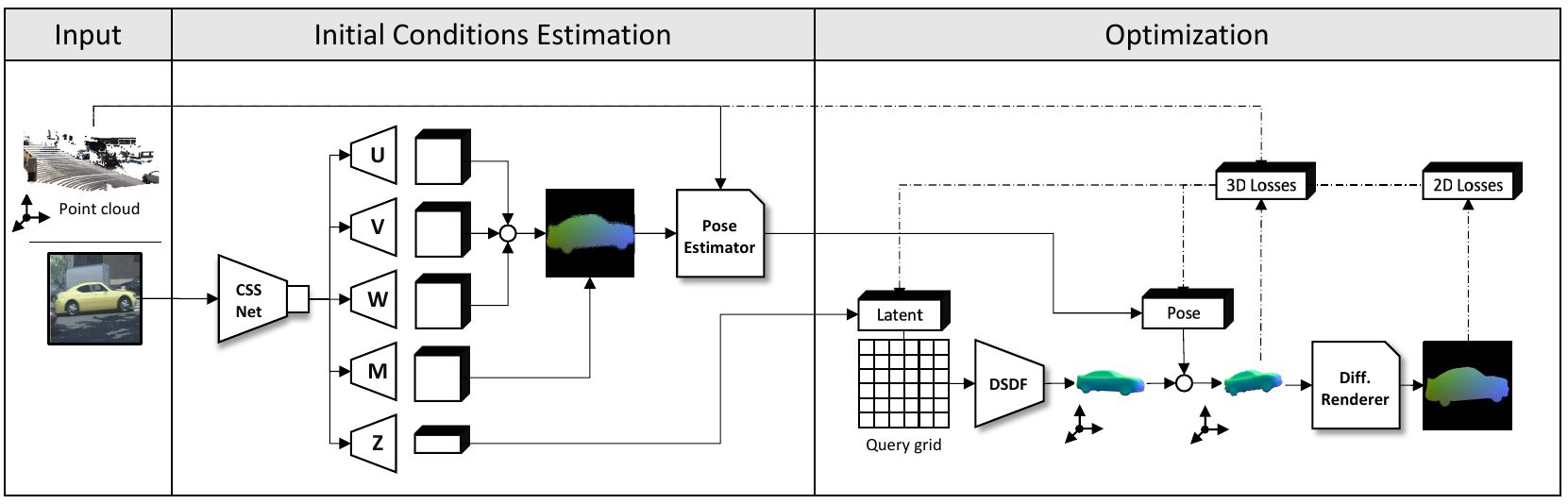}
	\caption{The detailed description of our annotation pipeline. We fetch frames from the dataset and separately process each 2D annotation with our CSS network. The CSS net outputs 2D NOCS for each RGB input pixel, mask, and a latent shape vector. The latent vector is fed to our surface projection module to get a normalized car point cloud and the initial pose. The estimated pose transformation is then applied to the car and the 3D losses are computed. Our differentiable renderer is used to define losses in the dense 2D screen space.}
	\label{fig:pipeline}  
\end{figure*}

\subsection{Optimization}
Given the output of the CSS network and our pose initialization, we proceed to the optimization stage (see Figure~\ref{fig:pipeline}). By concatenating the latent vector $\mathbf{z}$ with the query 3D grid $\mathbf{x}$ we form the input for our DeepSDF network. The DeepSDF net outputs SDF values for each query point on the grid, which we use for the 0-isosurface projection, providing us with a dense surface point cloud. The resulting point cloud is then transformed using the estimated pose and scale coming from the pose estimation module. The points that should not be visible from the given camera view are filtered using simple back-face culling, since surface normals have been already computed for 0-isosurface projection. At this stage, we are ready to apply 3D losses between the resulting transformed point cloud and the input LIDAR frustum points. The surface point cloud is also used as an input to our differentiable renderer, where we render NOCS as RGB and apply 2D losses between the CSS network's NOCS prediction and the renderer's output NOCS. The latent vector and the pose are then updated and the process is repeated until termination. 

3D losses allow us to get a precise pose/shape alignment with the frustum points. However, it is often the case that only few points are available resulting in poor alignment results. 2D losses, on the other hand, allow for precise alignment in the screen space over dense pixels, but are unsuitable for 3D scale and translation optimization, and heavily rely on their initial estimates. The combination of the two losses gives us the best of both worlds: dense 2D alignment and robust scale/translation estimation.

\section{Surface Tangent Discs}
Our surface tangent disc primitives formation requires solving a system of linear equations with a goal to compute the distance from the plane to each 2D pixel $(u, v)$:

\begin{equation}
    \begin{cases}
    u' = (u - o_u) \frac{z'}{f_u} \\
    v' = (v - o_v) \frac{z'}{f_v} \\
    Au' + Bv' + Cd - Au'_0 - Bv'_0 - Cd_0 = 0 \\
    \end{cases}
\end{equation}

The first 2 are the perspective projection equations and the third is a plane equation. If we solve the above system by a simple substitution, we get the following:

\begin{align}
\begin{split}
    d(\frac{A(u - o_u)}{f_u} + \frac{B(v - o_v)}{f_v} + C) \\ - Au'_0 - Bv'_0 - Cd_0 = 0 \longrightarrow \\
    d = \frac{Au'_0 + Bv'_0 + Cd_0}{(\frac{A(u' - o_u)}{f_u} + \frac{B(v' - o_v)}{f_v} + C)} \\ = \frac{n \cdot p_0}{n \cdot K^{-1} (u, v, 1)^T}
\end{split}
\end{align}

As a result, we can retrieve a 3D plane position $(u', v', d)$ for each 2D point $(u, v)$ on the screen and form primitives based on 3D distances from 3D shape points.


\end{document}